\documentclass[journal]{IEEEtran} 
\usepackage{times}
\usepackage{graphicx} 
\usepackage{subfigure} 
\usepackage{multirow}
\usepackage[cmex10]{amsmath}
\usepackage{amssymb}
\usepackage{algorithm}
\usepackage{algorithmic}
\usepackage{pgfplots}
\usepackage{adjustbox,subfigure,booktabs}
\usetikzlibrary{calc,positioning}
\usepackage[colorinlistoftodos,prependcaption]{todonotes}
\usepackage{blindtext}
\presetkeys%
    {todonotes}%
    {inline,backgroundcolor=cyan}{}
\tikzset{/tikz/notestyleraw/.append style={text=white}}
\usepackage{hyperref}

\begin{document}

\title{Constraint Selection in Metric Learning}
\author{\IEEEauthorblockA{Hoel Le Capitaine,\\}
\IEEEauthorblockA{Ecole Polytechnique de Nantes,\\ LINA UMR CNRS 6241,
            C. Pauc, Nantes, France\\Email : hoel.lecapitaine@univ-nantes.fr}}

\maketitle

\begin{abstract}
A number of machine learning algorithms are using a metric, or a distance, in order to compare individuals. The Euclidean distance is usually employed, but it may be more efficient to learn a parametric distance such as Mahalanobis metric. Learning such a metric is a hot topic since more than ten years now, and a number of methods have been proposed to efficiently learn it. However, the nature of the problem makes it quite difficult for large scale data, as well as data for which classes overlap.
This paper presents a simple way of improving accuracy and scalability of any iterative metric learning algorithm, where constraints are obtained prior to the algorithm. The proposed approach relies on a loss-dependent weighted selection of constraints that are used for learning the metric.
Using the corresponding dedicated loss function, the method clearly allows to obtain better results than state-of-the-art methods, both in terms of accuracy and time complexity. Some experimental results on real world, and potentially large, datasets are demonstrating the effectiveness of our proposition.
\end{abstract}



%

\section{Introduction}

The concept of distance, and more generally of norm, as well as the notion of similarity, are essentials in machine learning, data mining and pattern recognition methods. In particular, observations are grouped together depending on this measure in clustering, or compared to prototypes in classification. However, it is also well known that these measures are highly dependent of the data distribution in the feature space \cite{tversky1977features}. Historically, methods that are taking this distribution, or this manifold, into account are unsupervised (i.e. no class labels are available). Their objective is to project the data into a new space (whose dimension may be lower, for dimensionality reduction, or potentially larger, through kernelization) in which usual machine learning methods are used. The first, most established and widely used of these methods is certainly the Principal Component Analysis. In this kind of approach, called \textit{manifold learning}, the objective is to preserve the geometric properties of the original feature space while decreasing its dimension so as to obtain a useful projection of the data in a lower dimensional manifold, refer to e.g. MDS, ISOMAP, LLE, SNE (see \cite{MPH09} and references therein), or more recently t-SNE \cite{tsne}, a Student-based variation of SNE. 

Thereafter, class label information have been used in order to guide this projection, particularly by focusing on easing the prediction task (see e.g. Fisher linear discriminant analysis and its variants). Here again, the objective is to project the data into a new space that is linear combination of the original features. 

More recently, researchers tried to directly learn the distance (or similarity) measure in the original feature space, without projection.\footnote{We will see that in fact, using a Mahalanobis distance is equivalent to perform a linear projection of the data, and then compute the Euclidean distance in this new space.}

\begin{figure}[h!]
\centering\includegraphics[scale=0.4]{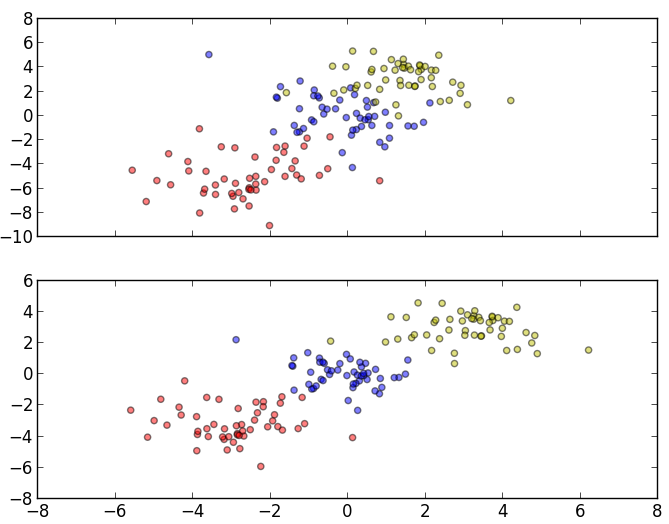}
\caption{\label{fig:ex}Initial data set where three classes (denoted as red, blue and yellow circles) are overlapping (up), and projected data set where discrimination is easier (bottom).}
\end{figure}
As opposed to manifold learning, which is unsupervised, \textit{metric learning} uses some background (or side) information.  For instance, in the seminal paper of Xing \textit{et al.} \cite{XNJ02}, the metric learning problem has been formulated as an optimization problem with constraints. The basic assumption behind this formulation is that the distance between similar objects should be smaller than the distance between different objects. Therefore, we generally consider whether pairs or triplets of observations as the constraints of the optimization problem.  More formally, given two observations $\mathbf{x}_i$ and $\mathbf{x}_j$ lying in $\mathbb{R}^p$, one wants to minimize the distance $d(\mathbf{x}_i, \mathbf{x}_j)$ if these two observations are considered as similar, and maximize the distance if they are considered as dissimilar \cite{XNJ02}. Alternatively, if the constraints are under the form of triplets of observations, we may minimize the distances $d(\mathbf{x}_i, \mathbf{x}_j)$ between similar objects and maximize the distances $d(\mathbf{x}_i, \mathbf{x}_k)$ between dissimilar objects, as in \cite{ITML}.

Depending on the application, this concept of similar and dissimilar objects can be obtained through their class labels for supervised problems, or with must-link and cannot-link, or side information for semi-supervised problems \cite{XNJ02}. This information can also be obtained interactively with the help of the user, where online learning algorithms are therefore well suited for this kind of application.

Due to the ever growing size of available data sets, online metric learning has also received a lot of interest. As opposed to batch learning where the entire learning set is available, online learning processes one observation (or pairs, triplets) at a time. Based on the output of each iteration, these approaches rely on getting a feedback of the quality of the metric (for instance a specified loss), and updating the parameters accordingly. This process is repeated until convergence of the metric. The results given by these methods are not as good as batch algorithms, but allows to tackle larger problems \cite{OASIS}.

The vast majority of metric learning approaches are using, as metric, the squared Mahalanobis distance between two $p$-dimensional objects $\mathbf{x}_i$ and $\mathbf{x}_j$ defined by
\begin{equation}
\label{eq:maha}
d^2_A(\mathbf{x}_i, \mathbf{x}_j) = (\mathbf{x}_i - \mathbf{x}_j)^T A (\mathbf{x}_i-\mathbf{x}_j)
\end{equation}
where $A$ is a $p$ $\times$ $p$ positive semi-definite (PSD) matrix. Note that if $A = I$, $d^2_A$ reduces to the squared Euclidean distance. In this setting, the learning task consists in finding a matrix $A$ that is satisfying some given constraints. In order to ensure that (\ref{eq:maha}) defines a proper metric (i.e. a binary function holding the symmetry, triangle inequality and identity properties), $A$ must remain PSD. Note that in the following, we denote this distance as $d_A(\mathbf{x}_i, \mathbf{x}_j)$. Note also that the matrix A is often the inverse covariance matrix of the data $X = \{\mathbf{x}_1, \cdots, \mathbf{x}_n\}$.

In practice, it may be intractable, so that several solutions have been proposed. The first one consists in relaxing the metric constraints. For example, in \cite{OASIS}, the authors use a bilinear similarity function defined by $s(\mathbf{x}_i, \mathbf{x}_j) = \mathbf{x}_i^T A \mathbf{x}_j$. In this specific case, $A$ is not required to be PSD, so that optimization is facilitated. 

Another solution consists in considering a factorization of $A$ as $L^T L$. This decomposition presents two major advantages over $A$. First,  the PSD constraint on $A$ is ensured and second, one can project the data into lower dimensional spaces by using the projection matrix $L$. In particular, if the rank of the matrix $A$ is $k$, then the matrix $L \in \mathbb{R}^{k \times p}$ is used to project the data into a k-dimensional space ($k < p$). Indeed, some simple algebraic manipulations show that one can write $d^2_A(\mathbf{x}_i, \mathbf{x}_j)$ as $\|L\mathbf{x}_i - L\mathbf{x}_j\|^2$.
This projection, similarly to manifold learning, allows to better separate the data for a classification task, see an example in Figure \ref{fig:ex}. In this sense, metric learning can also be seen as a supervised dimensionality reduction technique.

One of the most important step in metric learning is to define the constraints with respect to the available information (class labels, relative constraints). The vast majority of learning algorithms are choosing the constraints by randomly selecting pairs (or triplets) of observations that satisfy the constraint, and then feed this pair (triplet) as a constraint into the learning process. However, such random selection have several drawbacks. First it may not focus on the most important regions of the feature space (e.g. boundary of the classes), and second it remains constant over time, without taking into account the current metric.

In this paper, we propose to dynamically set the weight of constraints as a function of the current metric. In particular, the importance of less satisfied constraints (controlled by a margin) are up-weighted, and well satisfied constraints are down-weighted. Such constraints selection allows to focus on difficult observations of the feature space (often lying at the boundaries of classes).
Note that the proposed approach is not restricted to a particular metric learning algorithm. More precisely, it can be used in any iterative metric learning algorithm.
Before presenting in detail the proposed approach let us briefly describe some existing metric learning algorithms.

\section{Metric learning and related works}
The literature on metric learning is continuously growing, so that the presentation given here only mentions the most common and well known methods. The interested reader can refer to surveys on this topic, see e.g. \cite{SURVEY,kulis2012metric,bellet2015metric}. The general formulation of metric learning is to find $A$ such that $\ell(A, \mathcal{C}) + \lambda R(A)$, where $\ell$ is a loss function penalizing unsatisfied constraints, with $\mathcal{C}$ is the set of constraints. $\lambda$ is a trade-off parameter between regularization and the loss, and $R(A)$ is regularizer. This model is generally casted as a constraint optimization problem
\begin{align*}
\min &R(A)\\
\text{s.t. }& \ell(A,i) \leq 0, \forall i \in \mathcal{C}&
\end{align*}

Large Margin Nearest Neighbors (LMNN, \cite{LMNN}) is one of the early attempts to learn a Mahalanobis distance metric as a convex optimization problem over the set of PSD matrices.
The loss function is composed of the linear combination of two terms $\varepsilon_{pull}$ and $\varepsilon_{push}$. The first term aims at penalizing large distances between an observation and other observations sharing the same label, while the second term objective is to penalize small distances between observations from different classes. The loss function is then casted as a semidefinite program. Note that there no regularization term in the objective function so that LMNN is prone to overfitting. Another well known problem, related to the proposed approach, is that the selection of neighbors is initially made using Euclidean distance, which may not be adapted.

Information-Theoretic Metric Learning (ITML, \cite{ITML}, \cite{jain2012metric}) formulates the distance learning problem as that of minimizing the differential relative divergence between two multivariate Gaussian distribution under constraints on the distance function. In this approach, the regularizer $R(A)$ is taken as the LogDet divergence between successive $A_t$. The main benefit is that it ensures that $A$ remains positive semidefinite. Constraints are incorporated through slack variables, and the optimal matrix $A$ is obtained by successive Bregman projections.

It is related to an information-theoretic approach, which mean that there exists a simple bijection (up to a scaling function) between the set of Mahalanobis distances and the set of equal mean multivariate Gaussian distributions.

This method can handle general pairwise constraints, meaning it is sufficiently flexible to support a variety of constraints. ITML does not require eigenvalue computation or semi-definite programming, which allows it to be both efficient and fast for many problems. However, the computational complexity of updating in this algorithm is $O(cp^2)$, the cost increases as the square of the dimensionality. Therefore, this method is not suitable, at least in place, for large-dimensional  datasets.

The Online Algorithm for Scalable Image Similarity (OASIS) \cite{OASIS} is an online dual approach using the Passive-Aggressive \cite{crammer2006online} family of learning algorithms. It learns a similarity function with a large margin criterion and an efficient hinge loss cost. Its goal is to learn a parameterized similarity function of the form $S_W(\mathbf{x}_i, \mathbf{x}_j) = \mathbf{x}_i^T W \mathbf{x}_j$. With this formulation, $W$ plays a similar role as $A$ in metric learning. As in the metric learning formulation, the objective function can be written as the sum of a regularizer on $W$, the Frobenius norm, and a soft margin on the loss function.

OASIS uses a triplet $(\mathbf{x}_i, \mathbf{x}_j, \mathbf{x}_k)$  as the input for each iteration, meaning that class labels are unnecessary. More precisely, only an order of preference is needed (e.g. $\mathbf{x}_i$ closer to $\mathbf{x}_j$ than $\mathbf{x}_k$). In the proposed approach, there is no need for positive or symmetry constraints on the learned matrix. In particular, the bilinear similarity may not satisfy triangular inequality property. Removing this constraint allows to obtain a faster algorithm than when imposing a PSD metric matrix. 
Moreover, experiments show that iterative learning converges to a matrix $W$ that is close to symmetry.

Some other propositions adopt different approaches. For example, Maximally Collapsing Metric Learning algorithm (MCML, \cite{MCML})  relies on the simple geometric intuition that an ideal approximation of the equivalence relation is that all points of the same class should be mapped into a single location in the feature space, and points belonging to different classes should be mapped to other locations. In the training process, each time a new class or point is added to the training dataset, the distance between each training point and all the other points must be calculated. This leads to considerable computational complexity. 

To deal with this problem, Mensink et al. developed a class-independent method (NCM, \cite{NCM}) that has an almost zero cost when a new class is added. The NCM model is very similar to MCML, as it defines a distribution and makes this as close to ideal as possible. However, when defining the distribution, NCM only uses the distances between the point and each class mean. Therefore, new classes and training points can be added at near zero cost, with the drawback of loosing some local information for each sample.



A proposition, given in~\cite{Ebert12}, is more similar to our proposition, while presenting some differences. In their paper, the authors propose to use an exploitation-exploration criteria based on posterior probabilities. Therefore they impose the use of such a classifier, or at least an estimation posterior probabilities. In this work, the enhancement can be used in any stochastic metric learning algorithm, without any restriction on the kind of classifier being used. Moreover, the application they have in mind is to classify data, so that the criterion used focus on the uncertainty of the results provided by the classifier. In this paper, we focus  on the direct loss suffered from using only the distance, whatever the application (classification is a possible one, among others).

Finally, the authors, in \cite{SKW12}, propose a boosting-like algorithm for metric learning. While the idea of using the principle of boosting is similar to our proposition, they are using it to decompose the learning process to a linear positive combination of rank-one matrices, instead of setting local weights to individuals as in our approach. Consequently, the two methods do not focus on the same points, and may even be used together to produce an even more fast metric learning algorithm. 

In \cite{wang2007learning} and \cite{kar2011similarity}, the authors, relying on the theory of learning with good (dis-)similarity functions \cite{balcan2008theory}, propose some other heuristics to select examples. In particular, in \cite{wang2007learning}, their approach relies on AdaBoost, where several learners are trained on different subsets. In \cite{kar2011similarity}, the authors propose a selection that is promoting diversity. In practice, sample points are selected so that the average similarity between samples belonging to different classes is small. In other terms, it is maximizing the between-class variance of a subset of the data. In both cases, the objective is to provide a classifier, so that it is not the same objective as our proposition.

Our approach also differs from usual instance selection for nearest neighbor algorithms \cite{derrac2012integrating}, where a weight is given to the distance between two samples, whereas we propose to set a weight on samples.

In \cite{liu2012robust}, the authors also consider that points near the boundaries are important in order to obtain a good metric. More precisely, for each observation $\mathbf{x}_i$, the $a$ most different observations $\{\mathbf{x}_j\}_{j=1,a}$ of the same group, and the $k$ most similar observations $\{\mathbf{x}_k\}_{k=1,a}$ of a different group are used to build the constraint triplet $(i,j,k)$. However, this selection occurs only once, at the beginning of the process, and do not change during the learning of the metric. Consequently, this choice of triplets may be particularly unadapted to the data (since Euclidean is used without any prior). This approach is used in the experimental part, and is called Liu \& Vemuri.

In \cite{mei2013logdet}, Mei et al., starting from the same observations, propose to update the triplets at each iteration, as in our proposition. 
For each iteration, the performance of the current metric is estimated from the degree of disorder implied by this metric. More precisely, the number of time  distance between observations from different groups are lower than the distance between observations from the same group increases the disorder of the metric. Depending on this performance, triplets are dynamically selected so that points near boundaries are used as constraint points. This change of triplets over the iterations gives more interesting results than the triplet selection method proposed in \cite{liu2012metric}.
However, they only focus on points near boundaries; We propose to focus on points near boundaries, while also keeping also observations that are far from the boundaries. These observations are important because they allow to regularize the metric which would be too much over fitted if only boundaries are considered. Furthermore, their approach require the computation of the distance matrix between all sample pairs, which can be very time consuming, and even untractable for large data sets (as mentionned in \cite{mei2016learning}). This approach is used in the experimental part, and is called Mei et al.

Finally, note that we restrict here to the description of global metric learning algorithms: a unique metric matrix $A$ is learned for the entire data set. Some other approaches propose to learn a matrix by class \cite{LMNN} , or multi-task metrics \cite{parameswaran2010large}. Naturally, the proposition could also be used with non linear (i.e. kernelized or local metric learning, see e.g. \cite{wang2012parametric}, \cite{shi2014sparse}) methods.


This paper is organized as follows. We first present in section \ref{sec2} the problem of learning a metric in unevenly distributed feature spaces, and propose a way to choose the samples that will really help to learn this metric, in a fashion similar to active learning. Then we provide some experiments on some real world data sets in section \ref{sec3}. Finally, some comments and perspectives are drawn in section \ref{sec4}.

\section{Dynamic local weights for metric learning}
\label{sec2}
\subsection{Problem and proposed approach}
The problem of metric learning, when tackled by using a Mahalanobis distance, can be viewed as learning a linear projection of the data, and then compute a simple Euclidean distance in this new space. This approach gives better results than using the  Euclidean distance on the original space because of the distribution of the data. However, it generally needs heavy computation in cases where Euclidean distance could be sufficient. Indeed, the Euclidean distance gives satisfying results in regions where classes are well separated, and a specific treatment should be done only within the region of overlapping classes. In other words, a focus should  be given on difficult regions in terms of discrimination. The learned metric matrix should mostly depend on the individuals that are difficult to classify.
This can be done using various approaches, and we present in this paper a proposition related to instance selection, and the principle of boosting. The samples selected for the learning of the metric are the samples for which the classification is the hardest. 

Many propositions have been made on this topic, going from uncertainty based sampling, query by committee, to density based sampling, see e.g. \cite{settles2012active}. They can roughly be classified into two categories, whether they focus on exploration, or on exploitation.

In this paper, we propose to weight each incoming observation, and to update those weights for the next iteration according to the previous result, in an online fashion. More precisely, we are defining a loss corresponding to the residual error committed with the current metric matrix $A$. For each iteration, the loss is computed using the metric, and used for updating the weights of each observation. As mentioned earlier, the more the error, the more the probability to be selected during the next iteration.

There are different type of constraints in metric learning. They can be classified in the following four categories
\subsubsection{Class labels} 
for which only one object $\mathbf{x}_i$ is considered
\subsubsection{Pairwise labels} that generally implies that similar objects belong to the same class, whereas different objects belong to different classes. In practice, we define a set $\mathcal{S}$ of index pairs $(i,j)$ corresponding to similar objects $(\mathbf{x}_i, \mathbf{x}_j)$, and a set $\mathcal{D}$ of index pairs $(i,k)$ corresponding to dissimilar objects $(\mathbf{x}_i, \mathbf{x}_k)$. $\mathcal{S}$ and $\mathcal{D}$ are built using the class labels.
\subsubsection{Triplet labels} as its name indicate, involves three different objects $\mathbf{x}_i$, $\mathbf{x}_j$  and $\mathbf{x}_k$. In this scenario, the constraint is formulated as finding a pair $(i,j)$ in $\mathcal{S}$, and another pair, with same object $i$, $(i,k)$ in $\mathcal{D}$, the triplet being $(i,j,k)$. As in the pairwise case, $\mathcal{S}$ and $\mathcal{D}$ are built using the class labels. Expressed as a constraint, we have $d(\mathbf{x}_i,\mathbf{x}_j) \leq d(\mathbf{x}_i,\mathbf{x}_k)$.
\subsubsection{Relative labels\cite{liu2012metric}} are considering four different objects $\mathbf{x}_i$, $\mathbf{x}_j$, $\mathbf{x}_k$ and $\mathbf{x}_l$ for writing the constraint. In this setting, $\mathbf{x}_i$ and $\mathbf{x}_j$ are chosen so that their distance is lower than the distance between $\mathbf{x}_k$ and $\mathbf{x}_l$ : $d(\mathbf{x}_i,\mathbf{x}_j) \leq d(\mathbf{x}_k,\mathbf{x}_l)$. Contrary to pairwise or triplet, relative constraints are often used in modeling vague domain knowledge. Note however, that if $i=k$ or $j=k$, this approach reduces to a triplet one.

Naturally, for each of these approaches, the loss function should be different. In the sequel, we propose a loss function adapted to the different type of constraints.

Depending on the framework used by the learning algorithm, this loss may be a function of a pair of similar objects $(\mathbf{x}_i, \mathbf{x}_j) \in \mathcal{S}$, or a function of three objects $i,j, k$ for which we have $(i,j) \in \mathcal{S}$ and $(i,k) \in \mathcal{D}$.
For each of the possible type of constraints, we propose a corresponding loss function that can be used for setting the weight of each constraints during next iteration. 
Let us consider the case of pair-wise constraints. The associated constraint can be written using a bound $\gamma$ on the desired distance (that should be small), for instance, let say that $d_A(\mathbf{x}_i, \mathbf{x}_j) \leq \gamma$, because $\mathbf{x}_i$ and $\mathbf{x}_j$ are similar.

Therefore, we may now define the hinge loss incurred when using the current $A$ for $(\mathbf{x}_i)$, denoted $\ell_i$, as
\begin{equation}
\label{eq:losspairpos}
\ell_i = \max(0, d_A(\mathbf{x}_i, \mathbf{x}_j) - \gamma)
\end{equation}
Conversely, if $\mathbf{x}_i$ and $\mathbf{x}_k$ are in different groups, we define the corresponding loss as 
\begin{equation}
\label{eq:losspairneg}
\ell_i = \max(0, \gamma - d_A(\mathbf{x}_i, \mathbf{x}_k))
\end{equation}
In the second case, for triplet constraints, we may consider a combination of the two distances $d_A(\mathbf{x}_i, \mathbf{x}_j)$ and  $d_A(\mathbf{x}_i, \mathbf{x}_k)$. In particular, we set a margin between the two :  the latter must be larger than the former. In other terms, $d_A(\mathbf{x}_i, \mathbf{x}_k) - d_A(\mathbf{x}_i, \mathbf{x}_j) \geq \gamma$. Again, using an hinge loss formulation, we obtain
\begin{equation}
\label{eq:losstriplet}
\ell_i = \max(0, \gamma- d_A(\mathbf{x}_i, \mathbf{x}_k) + d_A(\mathbf{x}_i, \mathbf{x}_j) )
\end{equation}
Finally, for relative constraints, adding the margin $\gamma$ leads to consider the following loss function
\begin{equation}
\label{eq:lossrelative}
\ell_i = \max(0, \gamma- d_A(\mathbf{x}_k, \mathbf{x}_l) + d_A(\mathbf{x}_i, \mathbf{x}_j) )
\end{equation}
For all losses (\ref{eq:losspairpos}), (\ref{eq:losspairneg}), (\ref{eq:losstriplet}) and (\ref{eq:lossrelative}), if the constraints are satisfied, then the loss $\ell_i$ is zero, and is increasing as the difference between the actual and the expected distance is increasing. Note that we used a margin that is equal to 1. This may be also a parameter that one can estimate during the learning phase of the algorithm. In particular, depending on local information the required margin should be adapted. Note also that in this paper, we are using  a hinge loss, but other losses such as squared hinge, exponential loss, or logistic loss may also be used.

Without any prior on the distribution of the data, we uniformly draw the samples in the feature space. As the metric is being learned, it appears that attention should not focus on certain locations of the feature space, because discrimination in this part of this space is easy (in particular, far from the boundary between classes). On the contrary, in regions where classes are overlapping, learning the metric becomes harder. Therefore, we should select those hard samples more frequently, in order to learn the metric for these specific parts of the feature space. In order to do so, we propose to modify the way sample are selected when the metric is updated according to some incoming pairs or triplet of samples. More precisely, the new weight for the sample $\mathbf{x}_i$ during the next iteration $t+1$ is a function of the previously defined losses, and we define it as

\begin{equation}
\label{eq:upweight}
w_i^{t+1} = \frac{1}{Z^{t+1}}w_i^{t}\; e^{(\delta \alpha_i)}
\end{equation}
where $Z^{t+1}$ is a normalization factor ensuring that $\sum_i w_i^{t+1} = 1$, and $\alpha_i$ is defined, for normalization purpose on the error committed by $A$, as follows
\begin{equation}
\alpha_i = \frac{\ell_i}{d_A(\mathbf{x}_i,\bullet)},
\end{equation}
where $\ell_i$ is obtained using (\ref{eq:losspairpos}), (\ref{eq:losspairneg}), (\ref{eq:losstriplet}) or (\ref{eq:lossrelative}), and $\bullet$ is $\mathbf{x}_j$ for (\ref{eq:losspairpos}), and $\mathbf{x}_k$ for (\ref{eq:losspairneg}). For (\ref{eq:losstriplet}) and (\ref{eq:lossrelative}), $\bullet$ is also given by $\mathbf{x}_k$.
A large loss $\ell_i$, denoting that the constraints imposed for $\mathbf{x}_i$ are not respected, will increase the probability of it to be selected during the next iteration. Conversely, a null loss implies that the weight remains constant (up to the normalization). An additional parameter, $\delta$,  is controlling the rate change of the weights. A large value will converge very fast, but may miss some important points. In the following, this parameter is fixed using cross-validation. The corresponding algorithm is given in Algorithm~\ref{alg:example}. Note that without prior information, the initial metric matrix $A_0$ is the identity $I$, and the weights are uniformly distributed.
\begin{algorithm}[tb]
   \caption{Loss-dependent Weighted Instance Selection (LWIS) for Metric Learning\label{alg:example}}
   
\begin{algorithmic}
   \STATE {\bfseries Input:} data $X$ : $n\times p$ matrix,\\
   $A_0$ : the initial metric matrix,\\
   $\delta$ : weight rate,\\
   $\mathcal{S}$, $\mathcal{D}$ : similarity and dissimilarity constraints
   \STATE {\bfseries Output:} $A$ metric matrix
   \STATE $\cdot$ $A \gets A_0$,
   \STATE $\cdot$ draw $w$ as an uniform distribution on $n$,
   \REPEAT
   \STATE $\cdot$ randomly sample of $X$ according to $w$, to find whether\\
   \hskip5mm $\cdot$ a pair $(\mathbf{x}_i, \mathbf{x}_j)$ of similar objects,\\
   \hskip5mm $\cdot$ a pair $(\mathbf{x}_i, \mathbf{x}_k)$ of dissimilar objects,\\
   \hskip5mm $\cdot$ a triplet $(\mathbf{x}_i, \mathbf{x}_j, \mathbf{x}_k)$ such that $(i,j) \in \mathcal{S}$ and $(i,k) \in \mathcal{D}$,\\
   \hskip5mm $\cdot$ a quadruplet $(\mathbf{x}_i, \mathbf{x}_j, \mathbf{x}_k, \mathbf{x}_l)$ such that $d(\mathbf{x}_i, \mathbf{x}_j) < d(\mathbf{x}_k, \mathbf{x}_l)$\\ 
   \STATE $\cdot$ update $A$ according to the algorithm in use
   \STATE $\cdot$ compute the loss $\ell$ incurred with $A$ on the pair using (\ref{eq:losspairpos}) and (\ref{eq:losspairneg}), or on triplet using (\ref{eq:losstriplet}), or relative using (\ref{eq:lossrelative})
   \STATE $\cdot$ update weights $w$ using (\ref{eq:upweight})
   \UNTIL{convergence of $A$}\\
   {\bfseries returns} $A$
\end{algorithmic}
\end{algorithm}
\subsection{A synthetic example}
\begin{figure*}
\includegraphics[scale=0.26]{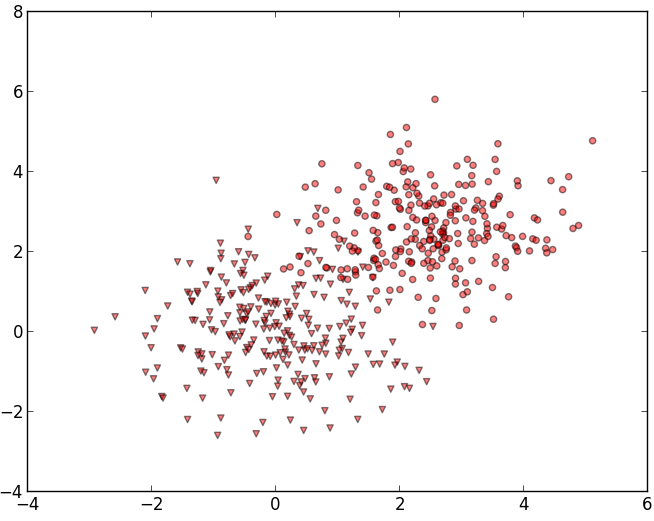}
\includegraphics[scale=0.261]{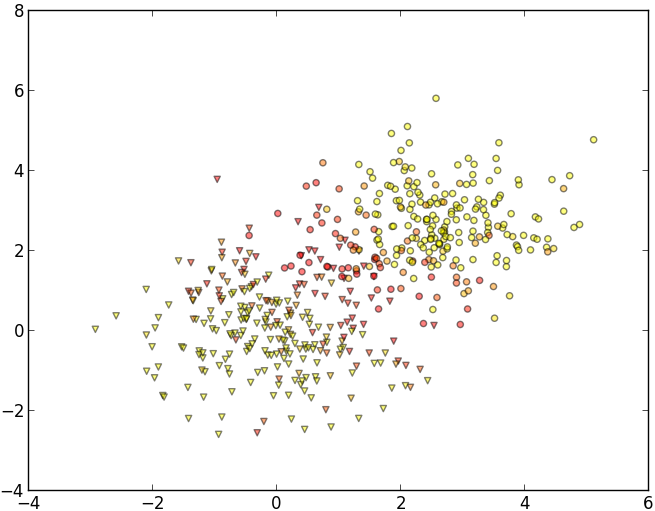}
\includegraphics[scale=0.26]{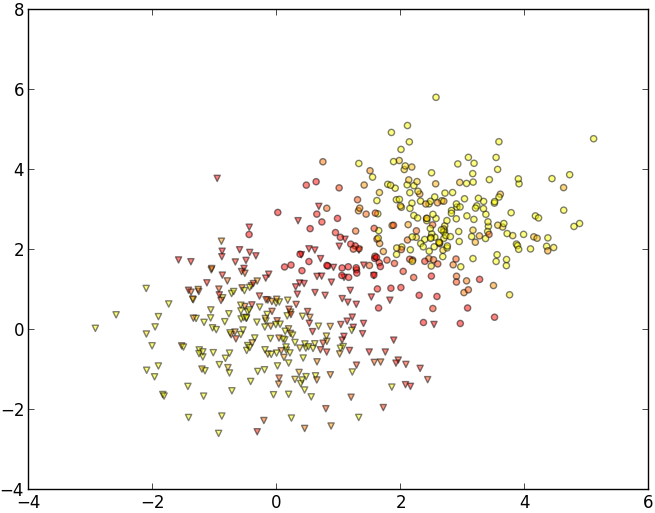}
\caption{A two-class ($\circ$ and $\triangledown$) toy dataset, and the evolution of individual weights over iterations : initialization (left), 50 iterations (middle) and 100 iterations (right). Color corresponds to the weight, from 0 (yellow) to 1 (red). As can be seen, the weights of samples lying near the separation of the two classes are increasing over time, while the other are decreasing.\label{fig:synth}}
\end{figure*}

In this section, we give an illustrative example on an synthetic toy dataset. The dataset is composed of two classes ($\triangledown$ and $\circ$), each following a two-dimensional normal distribution, that are slightly overlapping in the feature space, see Fig.~\ref{fig:synth}. Weights of each individual are encoded as a color, ranging from yellow (low weight) to red (large weight). The three plots correspond to the weights prior to the algorithm, after 50 iterations, and after 100 iterations, respectively. In this example, the weight rate $\delta$ is set to 0.1, and the margin $\gamma$ is set to 1.
As expected, weights are converging quite fast so that weights are large (i.e. red) in overlapping areas (where one wants to focus the attention of the learning algorithm), and low (i.e. yellow) in regions where discrimination is easy (i.e. the metric matrix is not discriminant for those individuals, and using $A$ as identity is sufficient to obtain good results). Note that weights do not differ that much between 50 and 100 iterations, meaning that convergence is fast for this simple case.


\section{Experiments}
\label{sec3}
\subsection{Experimental setup}

In this section, we provide some experiments on various datasets. We present some results on different data sets coming from the UCI repository~\cite{UCI}. In particular, we are considering the data set with different characteristics of dimension, volume and distributions. Numerical details are given in the Table \ref{tab:data}.
\begin{table}
\center
\normalsize
\begin{tabular}{lccc}
\toprule
dataset&\#features&\#observations&\#classes\\
\midrule
Balance&4&625&3\\
Wine&12&178&3\\
Iris&4&150&3\\
Ionosphere&34&351&2\\
Seeds&7&210&3\\
PenDigits&64&1797&10\\
Sonar&60&208&2\\
Breast Cancer&30&569&2\\

\bottomrule
\end{tabular}\vskip1mm
\caption{The eight data sets used for experimental evaluation and a summary of their main statistics.\label{tab:data}}
\end{table}

For comparison purpose, we consider the popular metric learning algorithm ITML~\cite{ITML}. We do not provide the results of using the Euclidean distance and using the inverse of the covariance matrix of the data for $A$, because they have already been unfavorably compared to ITML so that results gain clarity without these results. Moreover, the aim of this experiment is to compare the constraint (here triplet) selection method.
Naturally, the solution proposed in this paper can be applied to other metric learning approaches using random sampling during the learning of the projection matrix $L$ (or directly the positive semi-definite matrix $A$), e.g.~\cite{reg-online}). Depending on the application, the metrics of evaluation are different. In this paper, we consider the task of classification, where we are only interested in the most similar object(s) of the query. 

If one wants to perform the task of ranking individuals, then it should consider the $k$ most similar objects, and evaluate their consistency with respect to the query with usual information retrieval metrics (e.g. precision, recall, F-measure). In all experiments, the weight learning rate $\delta$ is set to 1, and the margin $\gamma$ is set to 2. A study of the influence of the parameters $\delta$ and $\gamma$ is provided at the end of the numerical experiments.  

In this experiment, we are comparing the metric obtained by the standard ITML algorithm (denoted as \textit{Random}), and how it can be improved by our proposition, noted as LWIS. In particular, we are using Algorithm 1, where the update of $A$ is given by the ITML procedure. More precisely, for all selection methods, the matrix $A$ is updated by
\begin{equation}
A_{t+1} = A_t + \beta A_t (\mathbf{x}_i- \mathbf{x}_j)(\mathbf{x}_i- \mathbf{x}_j)^T A_t,
\end{equation}
where $(i,j)$ are the index of the constraint observations, and $\beta$ is a projection parameter computed internally (see \cite{jain2012metric} for details).
Note that this algorithm takes as input pairs of observations that are similar ($\mathcal{S}$) or dissimilar ($\mathcal{D}$), so that it can be referred to a pairwise label constraint. In order to adapt this algorithm to generated triplet constraints, we adopt the following principle. From a given triplet $(i,j,k)$, we easily obtain two pairs $(i,j) \in \mathcal{S}$ and $(i,k) \in \mathcal{D}$. The two other triplet selection methods that we consider are the one proposed in \cite{liu2012robust}, noted \textit{Liu \& Vemuri}, and the one proposed in \cite{mei2013logdet}, noted \textit{Mei et al}. We stress out that we are not evaluating the metric learning algorithm, but the impact of constraint selection when using these methods, with the baseline metric learning algorithm ITML.
For each experiment, the number of constraints varies from 50 to approximately twice the number of individuals of the data, as it is commonly done in such evaluation, see e.g. \cite{kumar2008semisupervised} and \cite{liu2012metric}. 

\begin{figure*}
\centering
\includegraphics[scale=0.45]{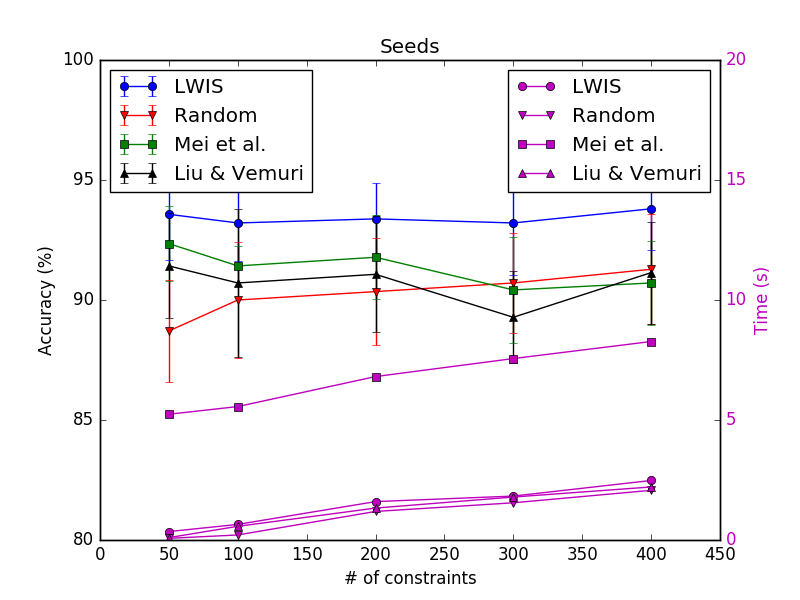}
\includegraphics[scale=0.45]{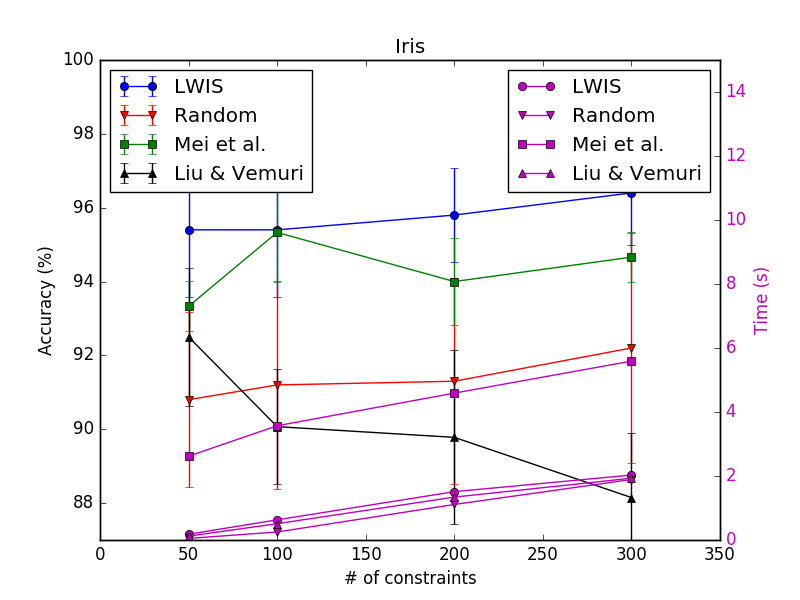}
\includegraphics[scale=0.45]{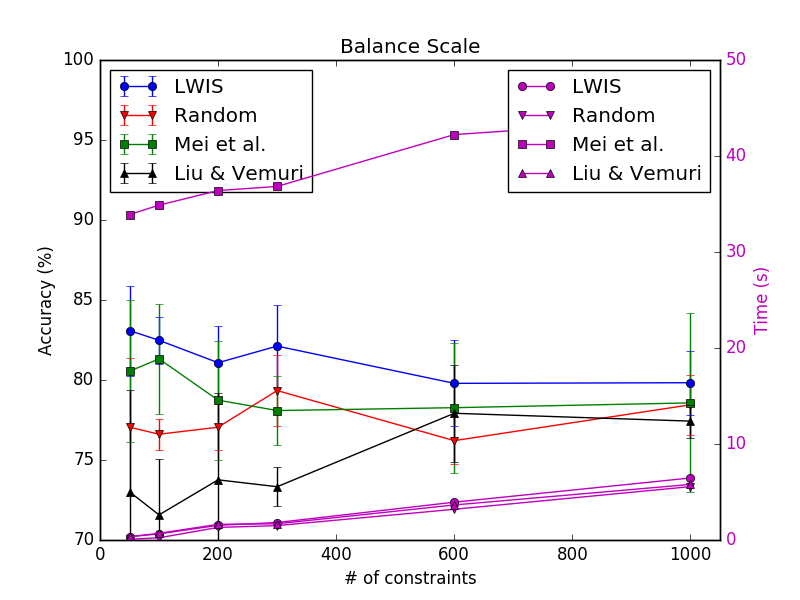}
\includegraphics[scale=0.45]{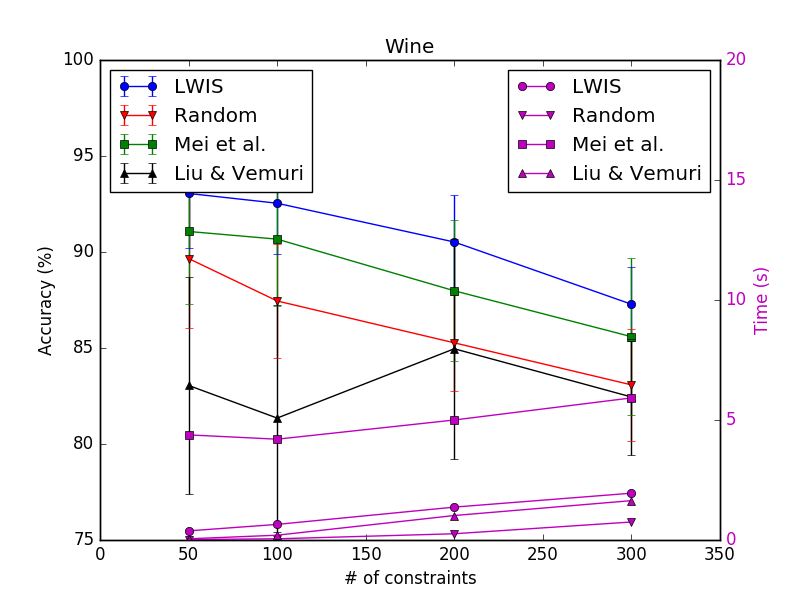}
\caption{Results for 4 datasets : Seeds, Iris, Balance Scale and Wine, from left to right. Time values (in magenta, right axis) do not use the same axis as accuracy (left axis).\label{fig:res1}}
\end{figure*}

\begin{figure*}
\centering
\includegraphics[scale=0.45]{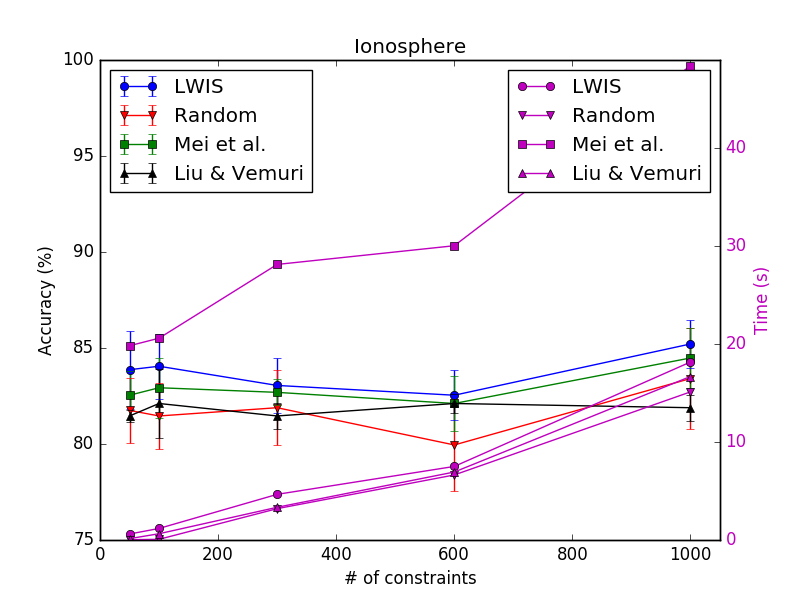}
\includegraphics[scale=0.45]{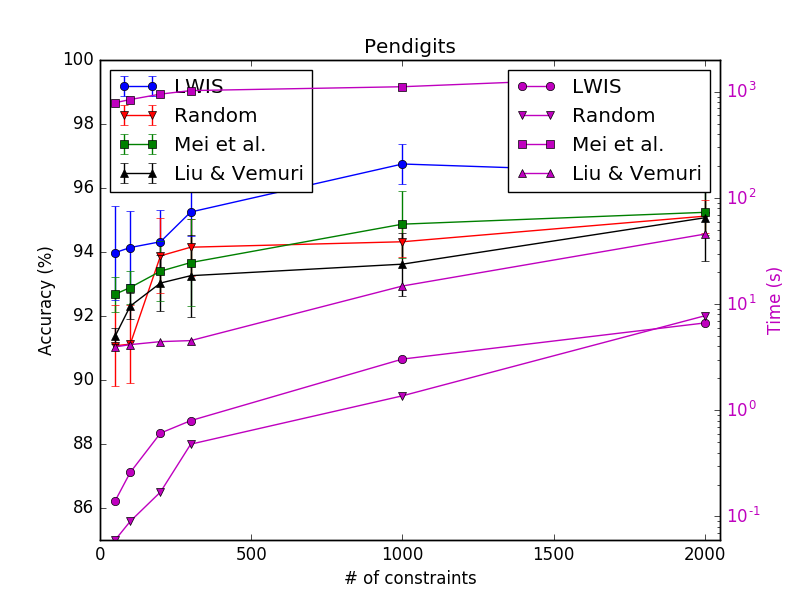}
\includegraphics[scale=0.45]{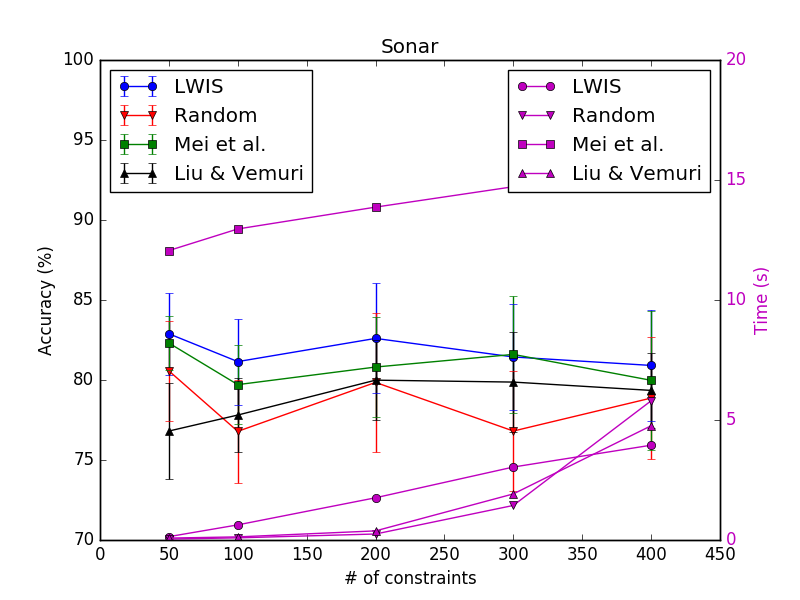}
\includegraphics[scale=0.45]{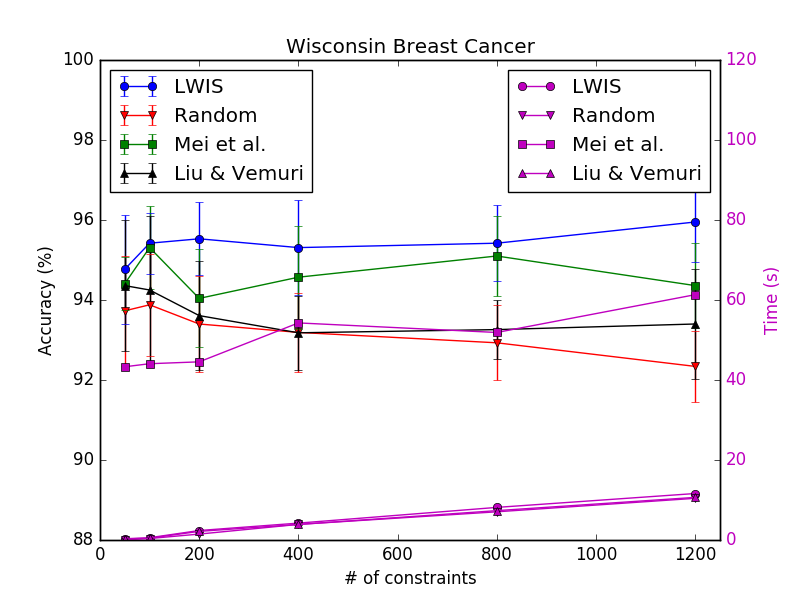}
\caption{Results for 4 datasets : Ionosphere, Pendigits, Sonar and Wisconsin Breast Cancer, from left to right. Time values (in magenta, right axis) do not use the same axis as accuracy (left axis).\label{fig:res2}}
\end{figure*}

The four approaches are then compared with two-fold cross validation, with k = 3 for k-NN classification.  Each experiment consists of 10 runs, and the mean value is indicated for each number of constraints, as well as the confidence interval at the 95 \% level for the mean value. Running times are also indicated for each data set in the same plot, in magenta color. Results are given in Figure \ref{fig:res1} and Figure \ref{fig:res2}.

The code for the experiments has been implemented in Python, and test were run on a Intel Core i7, 2.7 GHZ CPU, with 8 GB RAM.

\subsection{Numerical results}

The results of this first experiment clearly indicate that our method consistently results in a better predictive accuracy than the three other approaches  for all the considered data sets. Moreover, the running (learning) time of LWIS is better than the running time of Mei et al and Liu \& Vemuri. Without any inspection during the iterations, the random selection method is clearly faster than the three other approaches for all data sets. This can be easily explained by the fact the constraints are randomly selected \textit{prior} to the algorithm. The same behavior appears for the method of Liu \& Vemuri. Constraints are selected prior to the algorithm, according to the position of the observations in the feature space. Therefore, during the learning process, no time is lost for considering new weights. This is naturally not the case of our method and the method proposed by Mei et al. The consequence is that the performance of these two approaches clearly outperform the two others. The price of this performance for Mei et al is that it computes the distance matrix in each iteration, therefore resulting in potentially very long running times (see e.g. the PenDigits datasets, where the number of observations is moderately large, and see next subsection for a detail analysis on a large scale problem). Our method also change the weights at each iteration, but the cost is very moderate, since it only compute the distances implied in the current constraint, using the current metric $A_t$. Consequently, the running time of our method is similar to \textit{prior selection} methods (Random and Liu \& Vemuri).

During the experiments, it also appeared that prior selection methods (Random and Liu \& Vemuri) tends to converge in a fewer number of iterations than \textit{online selection} methods (Mei et al. and our method), and so particularly when the number of constraints is low. This can be explained by the fact that having constant constraints implies that the same information is used again and again, until the metric does not change anymore. On the other hand, changing the constraints very often does not facilitate the convergence of the algorithm. Nonetheless, the method proposed by Liu \& Vemuri, takes often more time than our method, and particularly when the size of the data increases. For instance, the running times are almost all the same, except for the PenDigits data set, for which the number of observations is equal to 1797. In this case, Liu \& Vemuri method clearly takes more time, due to the distance matrix computation prior to the algorithm.

It is interesting to point out that the predictive accuracy is increasing as the number of given constraints is also increasing for all data sets except Balance-Scale and Wine. Finally, one can note that the variance of the accuracy tends to decrease as the number of constraints increases, whatever the selection method.

In order to compare multiple constraint selection methods over multiple datasets, a combination of a Friedman test and a Nemenyi post-hoc test is used, following the recommendations of \cite{TEST}. Let $R_j^i$ be the rank of the $j$-th method on the $i$-th dataset. The Friedman test compares the average ranks $R_j$ over all datasets. Under the null-hypothesis, stating that two similarity measures are equivalent, their ranks should be equal (here $R_{j} = 2.5$ for all $j$). The Friedman statistic is given by
\begin{equation}
\chi^2_F = \frac{12N}{ns(ns+1)} \left( \sum_j R^2_j - \frac{ns(ns+1)^2}{4}\right)
\end{equation}
where $N$, the number of datasets, and $ns$ the number of constraint selection methods, are big enough, typically $N>10$ and $ns>5$.

The Friedman test  proves that the average ranks are significantly different from the mean Rank $R_j = 2.5$ expected under the null hypothesis. The corresponding $p$-value is almost zero, so that the null hypothesis is rejected at a high level of confidence.

\begin{figure}
\includegraphics[scale=0.46]{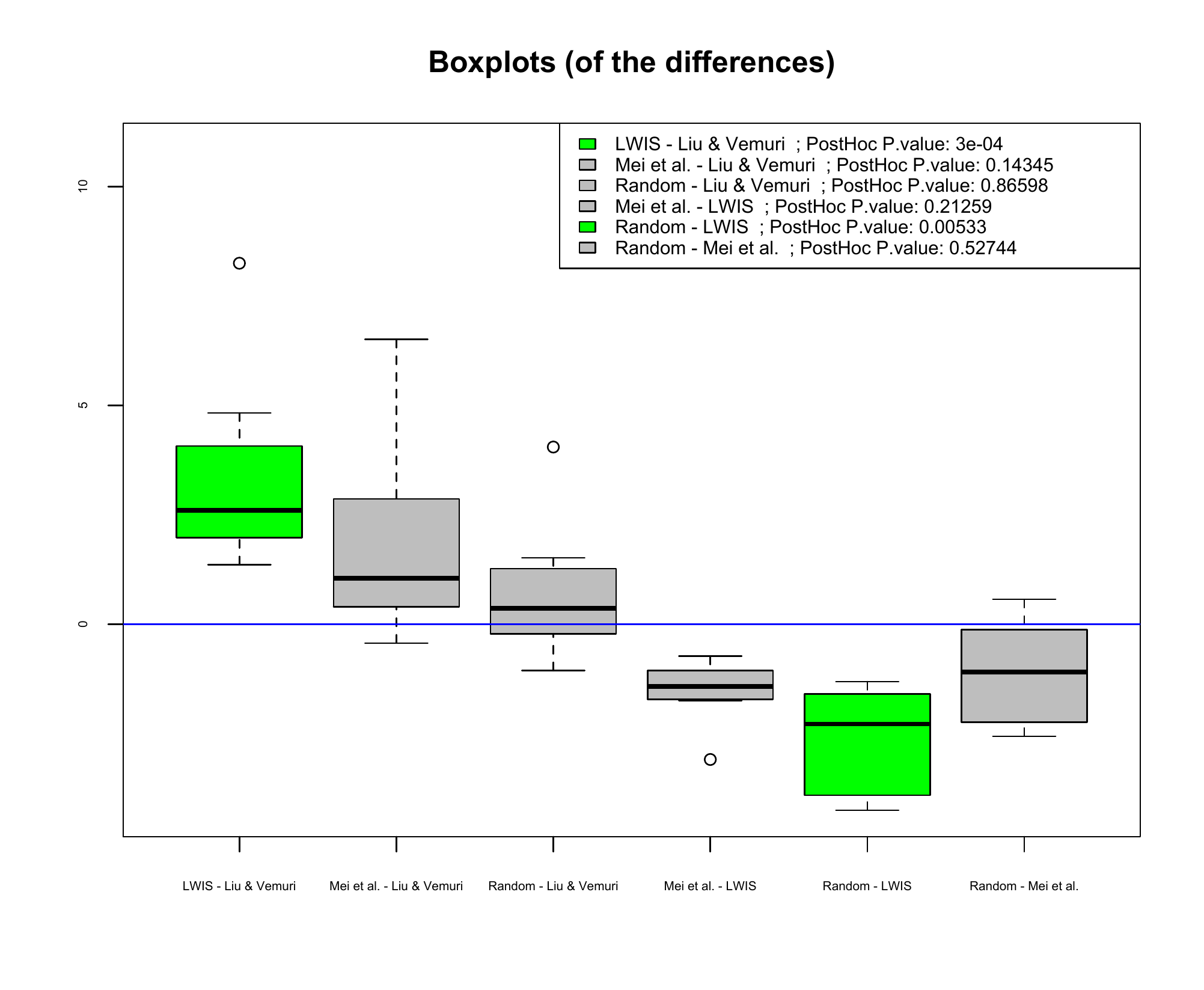}
\caption{Results of the Friedman test for all pairs of constraints selection methods\label{fig:boxplot}}
\end{figure}

If the null hypothesis is rejected,  the Nemenyi post-hoc test is proceeded. The performance of two similarity measures is significantly different if the corresponding average ranks differ by at least the critical difference, defined by
\begin{equation}
CD = q_\alpha \sqrt{\frac{nc(nc+1)}{6N}},
\end{equation}
where $q_\alpha$ values are based on the Studentized range statistic divided by $\sqrt{2}$, (see \cite{TEST} for details).
Figure \ref{fig:boxplot} illustrates the results of the Friedman test for all pairs of constraint selection methods. As can be seen, a statistically significant difference is observed for the couples (LWIS, Liu \& Vemuri) and (LWIS, Random) (in green).

In a second experiment, we study the influence of the margin $\gamma$ on the quality of the results. In this experiment, the number of constraints is set to the number of individuals of the data set, and the margin varies from 1 to 5.
Note that, due to the formulation of the constraints of ITML, we are using the loss defined by the Eq. (\ref{eq:losspairpos}) and (\ref{eq:losspairneg}) for the experiments. Corresponding results (accuracy) are given in Table \ref{tab:res2} (without confidence interval for clarity). For consistency, each run is repeated ten times, and mean accuracy is reported in the table. Bold values indicate the best performance obtained for each dataset.
For both parameter $\delta$ and $\gamma$, the difference of running time when changing the parameter is not statistically significant, so that it is not plotted for clarity.
\begin{table}
\center
\normalsize
\begin{tabular}{lccccc}
\toprule
dataset&\multicolumn{5}{c}{Margin $\gamma$}\\
&1&2&3&4&5\\
\midrule
Balance Scale&78.84&81.06&\textbf{81.15}&79.61&78.84\\
Wine&89.83&91.18&90.84&91.52&\textbf{91.86}\\
Iris&95.2&\textbf{96}&95.6&94&95.75\\
Ionosphere&82.65&\textbf{84.48}&83.18&83.83&83.29\\
Seeds&90.53&91.42&\textbf{92.07}&91.07&90.76\\
PenDigits&96.50&\textbf{97.05}&96.25&96.73&96.42\\
Sonar&79.71&78.80&78.44&80.43&\textbf{80.97}\\
Breast Cancer&\textbf{95.34}&95.01&94.68&95.27&95.01\\
\midrule
mean & 88.56 & 89.375 & 89.06 & 89.05 & 89.11\\
\bottomrule
\end{tabular}\vskip1mm
\caption{Accuracy of LWIS as a function of the margin $\gamma$\label{tab:res2}}
\end{table}
From this table, one can see that setting $\gamma$ to 2 \textit{generally} gives good results (3 first rank out of 8), but having $\gamma = 3$ or $\gamma = 5$ also gives interesting results (2 first rank out of 8). In order to deeply analyze the statistical difference between the values of $\gamma$, we conduct a Friedman test, as recommended in \cite{TEST}.
The Friedman statistic of these results is equal to 5.962, which is lower than the critical value 9.200 at the level 5\%. Consequently, the null hypothesis, stating that all margins lead to the same performance, is not rejected. A change of margin is not statistically significant. No post-hoc tests are necessary in this case.

\begin{figure*}
\includegraphics[scale=0.29]{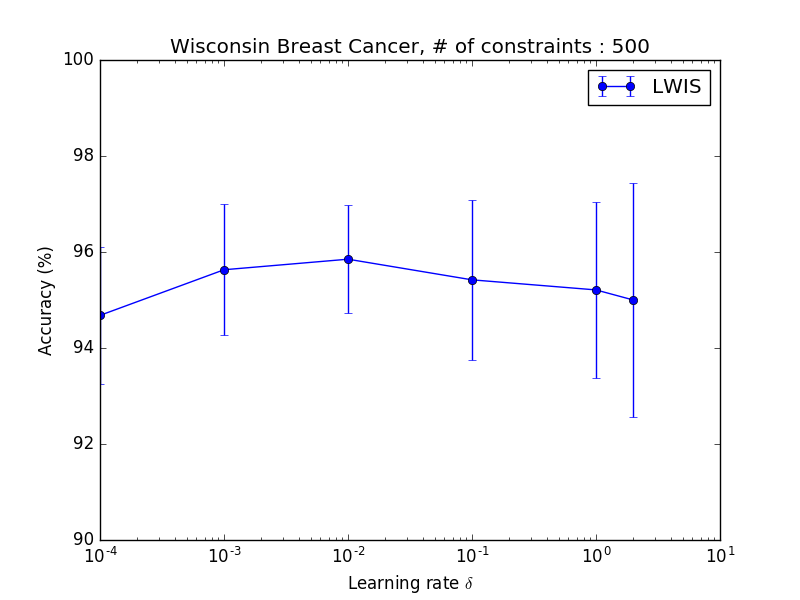}
\includegraphics[scale=0.29]{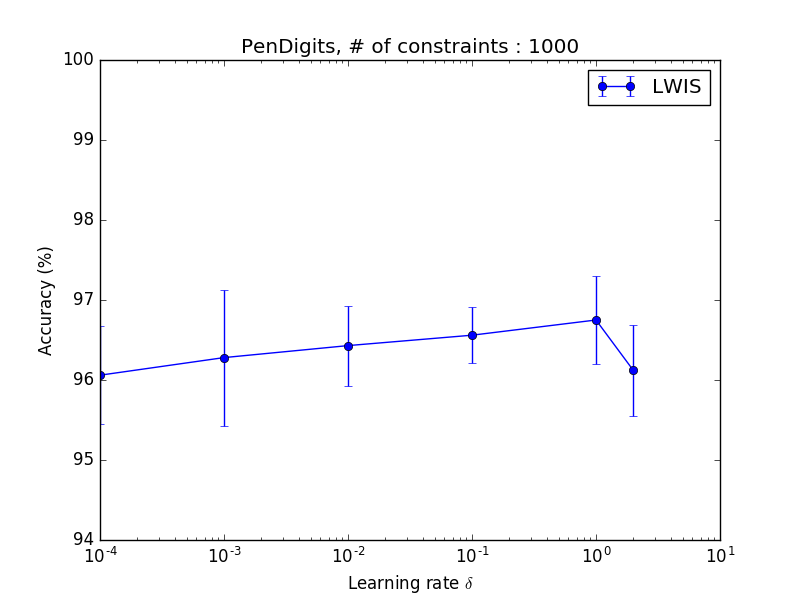}
\includegraphics[scale=0.29]{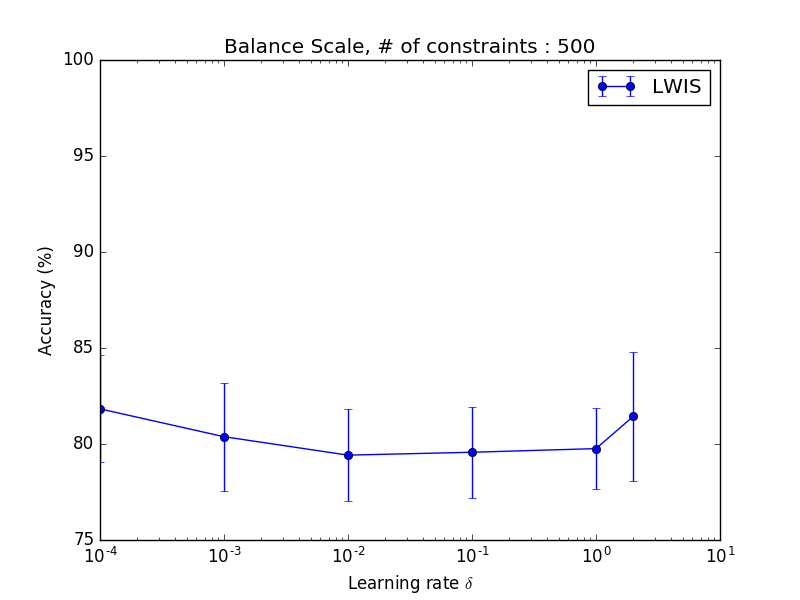}
\caption{An illustration of the relationship between accuracy and the learning rate $\delta$, ranging from $10^{-4}$ to $2$. Experimental results based on three datasets: Wisconsin Breast Cancer (left), PenDigits (middle) and Balance Scale(right). \label{fig:delta}}
\end{figure*}

The third experiment is about the study of the influence of the learning parameter $\delta$. For this experiment, we consider only three datasets (for brevity): Balance-Scale, PenDigits and Wisconsin Breast Cancer. For each of the datasets, the parameter $\delta$ belongs to one of the following values: $10^-4$, $10^-3$, $10^-2$, $10^-1$, $10^0$, $2$, and the corresponding average (over 10 runs) accuracy is reported. Corresponding results are given in Figure \ref{fig:delta}.
From Figure \ref{fig:delta}, one can draw the following comments. From the three figures, we can remark that there no $\delta$ value for which the maximum is reached, whatever the data set. In particular, this maximum is respectively reached for $\delta = 10^, {-2}$, $\delta=1$, and $\delta = 10^{-4}$. In the previous experiments, this parameter $\delta$ was constant, and equal to $\delta = 1$, but searching the value adapted to the data, through grid search, should improve the result of the method. Nonetheless, one can observe that the difference between maximum accuracy and minimum accuracy is not larger than 2\%, whatever the dataset.
\subsection{A large scale experiment : Forest cover type}
\begin{figure*}
\centering
\includegraphics[scale=0.45]{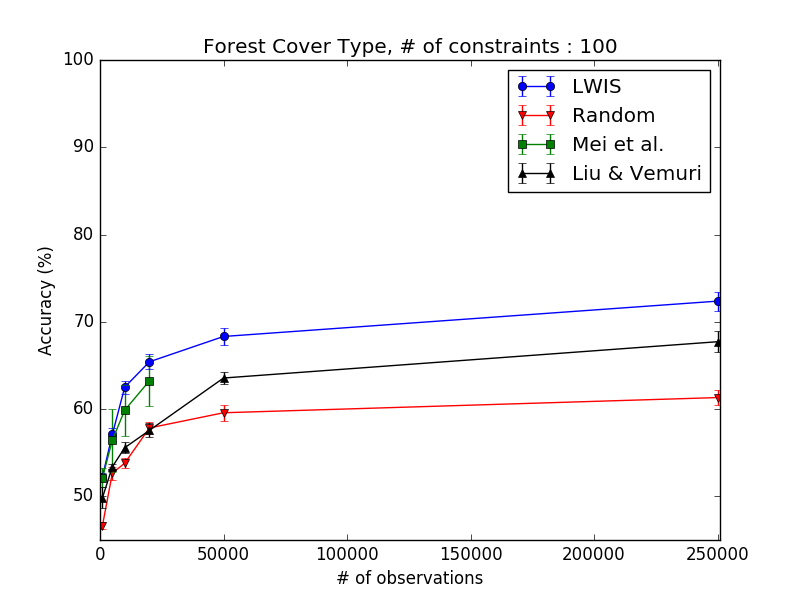}
\includegraphics[scale=0.45]{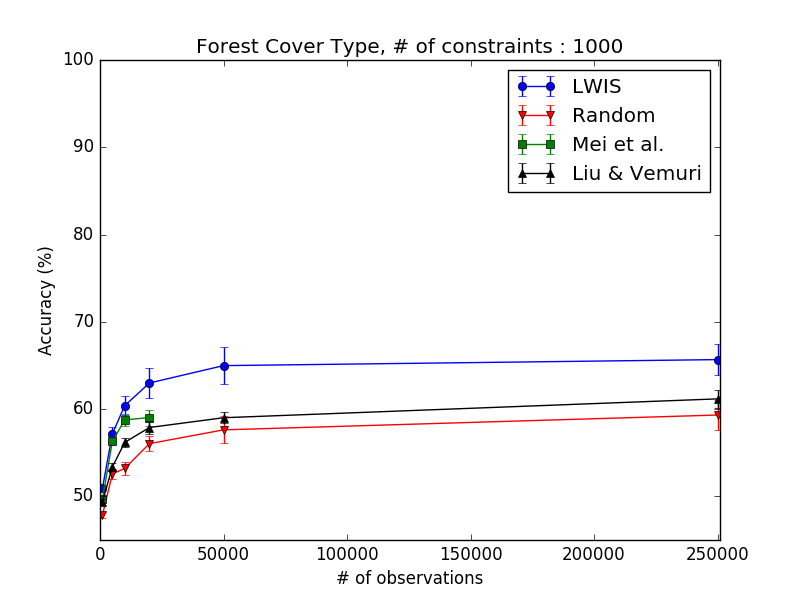}
\caption{Results of the two algorithms for the Forest Cover Type dataset. The number of constraints is set to 100 (left) and 1000 (right).\label{fig:res3}}
\end{figure*}

We now conduct an experimental study on a larger data set, both in terms of volume and dimension: Forest cover type. The observations are taken from four wilderness areas in Roosevelt National Forest. The number of observations is $n=581012$ Each observation is a 30 x 30 meter cell, described by $p=54$ geographical and physical features. The associate task is to predict the cover type of each cell, among the seven different cover types. It is freely available from the UCI machine learning repository \cite{UCI}. In metric learning, the metric is often defined by its associated matrix $A$. Since the dimension of $A$ is a function of the dimension of the data, this parameter is critical when it increases.
In this section, we studied how the running time of LWIS scales with the size of the training set.
In particular, the number of randomly selected observations  sampled from the initial 581,012 ones varies from 1,000 to the entire data set, so that we can inspect the impact of increasing the volume of the data.
The number of constraints also varies, for the same reasons as in the experiments of the previous subsection. A first run consists in using only 100 constraints, and then inspect the changes in a second run by using 1,000 constraints. The corresponding results are given in Figure \ref{fig:res3}, left and right, respectively.

Here again, whatever the number of observations and the number of constraints, LWIS performs better than the three other methods, closely followed by Mei et al. for a relatively small number of observations.
It shows the effectiveness of our method, even with large (in volume and dimension) data sets.
As can be observed, whatever the number of constraints, the accuracy increases as the number of observations increases, and reach its maximum when all observations are considered.
It appears that adding constraints for this data set does not improve the performance of nearest neighbor classification.
We give in Table \ref{tab:timeforest} the run times, in seconds, with a varying number of observations and constraints for the Forest Cover Type data set.
\begin{table}
\begin{tabular}{cccccc}
\toprule
Constraints & Observations & \multicolumn{4}{c}{Method}\\
&&LWIS&Random&Mei et al.& Liu \& Vemuri\\
\midrule
\multirow{7}{*}{100}& $n=1000$ &0.69& 0.18&84.88 &0.69\\
					& $n=5000$ & 1.03& 0.55&2124.38&11.44\\
					& $n=10,000$ & 1.87& 1.51&8670.17&47.43\\
					& $n=20,000$ & 10.40&5.95 &33568.96&178.31\\
					& $n=50,000$ & 21.67&13.89 &--&868.46\\
					& $n=250,000$ & 109.46& 89.90&--&20658.49\\
\midrule
\multirow{7}{*}{1000}& $n=1000$ &6.51& 5.60&100.32 &5.26\\
					& $n=5000$ & 6.23& 5.19&2441.93&20.22\\
					& $n=10,000$ & 8.62& 7.36&9694.46&62.44\\
					& $n=20,000$ & 15.89&11.74&38874.78&236.46\\
					& $n=50,000$ & 456.3&244.68 &--&912.73\\
					& $n=250,000$ & 2017.38&1258.07&--&21549.55\\
\bottomrule
\end{tabular}
\caption{Running times, in seconds, with a varying number of observations and constraints for the Forest Cover Type data set.\label{tab:timeforest}}
\end{table}

As can be seen, both Mei et al. and Liu \& Vemuri approaches scale quadratically with the number of observations, hence they are untractable for a large volume of data. As expected, the fastest method is still the random one, and the slowest the one that is computing the distance matrix in each iteration (Mei et al.).
Naturally, when increasing the number of constraints, the running time also increases, but the impact is not as important as one could expect.

\begin{figure*}
\centering
\includegraphics[scale=0.29]{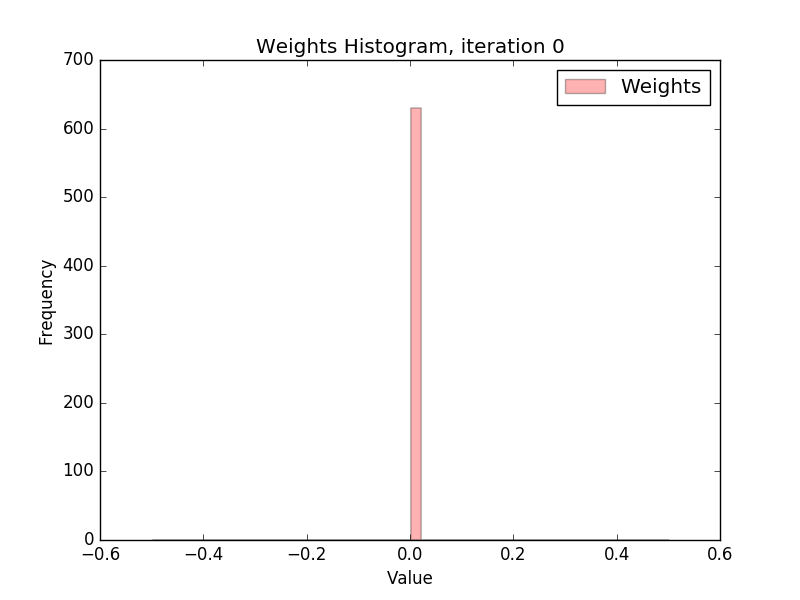}
\includegraphics[scale=0.29]{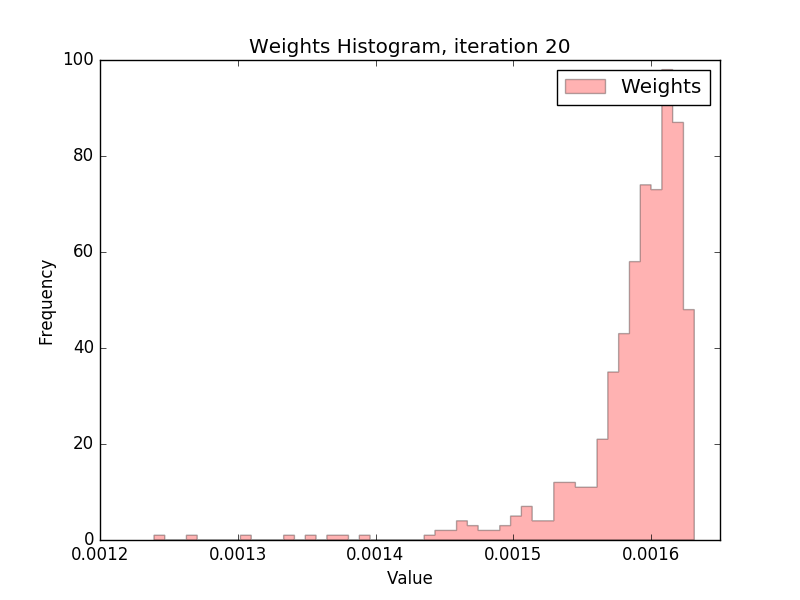}
\includegraphics[scale=0.29]{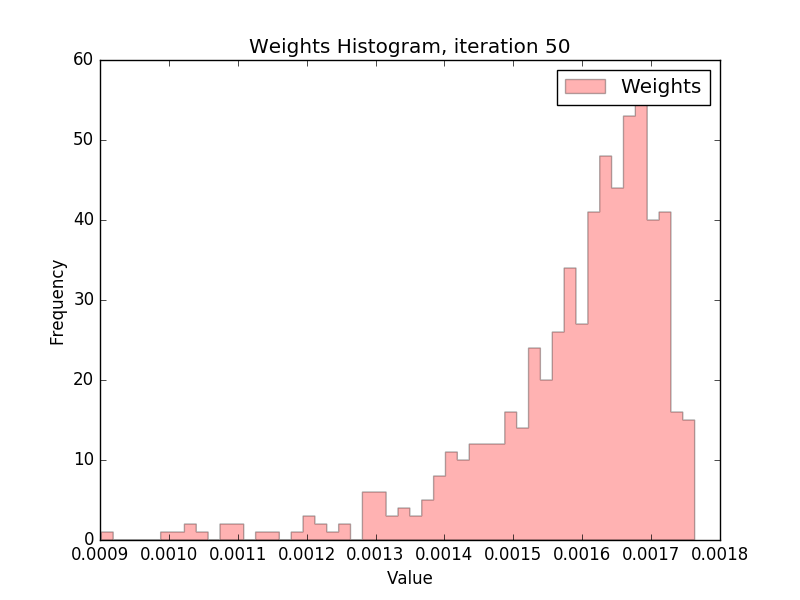}
\caption{Distribution (frequency) of the weights of constraints for the Forest Cover Type data set at different iterations. From left to right: initialization, after 20 iterations, and after 50 iterations.\label{fig:res4}}
\end{figure*}

In Figure~\ref{fig:res4}, the distribution of weights along 50 iterations are given, starting from the initialization (left), after 20 iterations (middle) and after 50 iterations (right). As can be observed, distribution starts from an uniform one, and quickly reach a bell-shaped distribution. On the last histogram, one can see that the majority of observations get a weight around 0.0016 and 0.0017, a few (near boundaries) points get larger weights, and the rest, considered as easy to classify, get low weights.
Note that there are no constraints with weight to zero, meaning that all the constraints can be potentially selected.

\section{Conclusion}
\label{sec4}
In this paper, we presented a simple way of improving accuracy and scalability of iterative metric learning algorithms, where constraints are obtained, or built, prior to the algorithm. The proposed approach relies on a loss-dependent weighted selection of constraints that are used for learning the metric. The loss depends on the current metric and evaluates the difficulty of classification for the considered observations. By using this dedicated loss function, the method clearly allows to obtain better results than state-of-the-art methods, both in terms of accuracy and time complexity. Compared to other difficulty-based constraint selection methods, the difference in time complexity is particularly important for large, in volume and dimension, data sets.


Among the perspectives we have in mind, let us mention the study of the convergence and stopping criterion of the method, by the help of works on early stopping in boosting, see e.g.~\cite{Zhang05}.
Furthermore, we want to work on the adaptation of this framework to online learning algorithms. In this version, we consider a batch version where constraints are given weights, but this is impossible for the online setting : constraints arrive one at a time. One first solution to this problem is to locally set the importance of an incoming constraints according to the density of loss within the neighborhoods of already considered data, or following some stream based sampling approaches \cite{roy2001toward}. Regret bounds can be easily retrieved in this context. 

Finally, in the conducted experiments, the size (in terms of volume) of the considered data sets is rather small, with the notable exception of Forest Cover Type. We would like to conduct some experiments and study the adaptability and the scalability of the proposed method to large scale problems (e.g. data where the number of individuals is larger than one million, and the number of features greater than one thousand). We will also provide a deeper study on other, more recent, metric learning algorithms that are using random constraint selection, e.g. \cite{liu2012metric}, \cite{qi2009efficient}. The same study can be conducted on random-based constraint selection for similarity learning algorithms, as in \cite{OASIS} and \cite{yao2015sparse}.




%

\bibliographystyle{IEEEtran}
\bibliography{bibliography}

\end{document}